# A Generative Model for Multi-Dialect Representation

[*]Emmanuel N. Osegi [1]

[1]Department of Information and Communication Technology, National Open University of Nigeria, Lagos State, Nigeria

## Abstract

*In the era of deep learning several unsupervised models have been developed to capture the key features in unlabeled handwritten data. Popular among them is the Restricted Boltzmann Machines (RBM). However, due to the novelty in handwritten multi-dialect data, the RBM may fail to generate an efficient representation. In this paper we propose a generative model – the Mode Synthesizing Machine (MSM) for on-line representation of real life handwritten multi-dialect language data. The MSM takes advantage of the hierarchical representation of the modes of a data distribution using a two-point error update to learn a sequence of representative multi-dialects in a generative way. Experiments were performed to evaluate the performance of the MSM over the RBM with the former attaining much lower error values than the latter on both independent and mixed data set.*

## 1. Introduction

Deep Learning can be traced to the works of Alexey Grigorevich Ivakhnenko appearing under the name Group Method of Data Handling (GMDH) and Fukushima's Neocognitron-a self-organizing unsupervised neural network for learning temporal patterns [1], [2]. It is a field that seeks to discover novel or hidden

[*]Corresponding author: Emmanuel N Osegi: National Open University of Nigeria, Information and Communication Technology Department, 14/16 Ahmadu Bello Way Victoria Island, Lagos Nigeria. E-mail: nd@osegi.com

.data features using multiple paths or layers, big data and various machine learning techniques with the hope of extracting features that best explains the data. More recently deep generative models such as the RBM originally developed as the "Harmonium" (an analytic RBM) in [22] has been developed further in [25], [3] classification RBM [23], [24] and conditional RBM [4]. The Restricted Boltzmann Machine (RBM) have been applied to handwritten recognition [5], Gaussian generative models in [6], and with careful modifications to word observations RBM's in [7]. However one key question that still remains unanswered is how good is a machine representation of a natural language – in this case handwritten dialects. Things become complicated when there is a mixture of words or dialects and simple generative models such as RBMs no longer becomes sufficient.

In this paper we make three important contributions. First, we develop a new generative model, the Mode Synthesizing Machine (MSM) for synthesizing in a hierarchical (columnar) fashion, the modes of a data distribution and learning their representations using a two-point error-update. Second we develop a technique for human representation of written natural language based on a write-the-way-you-hear (WYH) concept. Third, we make a modification to the Rectified Linear Units used as learning units in current deep learning systems and apply it to the MSM.

The remainder of this paper is structured as follows:

In Section 2 we discuss related works particularly as it relates to handwritten language patterns. In Section 3 we introduce the Representative learning based on the multi-dialect-transformation concept, the RBM's used in training current generative models for handwriting recognition, the modified Rectified Linear Unit (mReLU) and our proposed Mode Synthesizer Machine (MSM). In Section 4, we present our experiments and results. Finally, we give our conclusions in Section 5.

## 2. Related Literature

Handwritten character recognition from the representative point of view is a machine learning task devised to extract interesting features or patterns from a sequence of handwritten words or characters while using these features to recognize similar characters at a later point in time. Thus, it may be regarded as a predictive process. As a supportive task, it may be combined with other sub-tasks such as speech processing, video and image processing by way of special embeddings. Over the years, machine learning researchers have developed a variety of applications to meet diverse handwritten recognition needs. Handwritten recognition have been performed using a hierarchical products of experts (PoE) approach based on the Boltzmann machines [5], mixtures of linear models [9], and Gaussian generative models [6]. However, one drawback in using these models is the computational expense in feature processing largely attributed to the "curse of dimensionality". Such drawbacks may however, be overcome using multiple explanatory factors or manifold operators as in de-noising auto-encoders, see for example in [8]. One current approach as in [7] is to modify the Gibbs sampler using carefully designed Metropolis-Hastings transitions to reduce the dimensionality in data.

Notwithstanding the developments in generative models such as RBM's in handwritten recognition applications several other related applications have been employed in machine learning (ML) handwriting tasks. In [10] a deep convolutional neural network was developed for online handwritten Chinese character recognition. By incorporating domain specific features they reported improved recognition accuracies.

Dynamic Ensemble classifiers (DECS) employing a mixture of ML techniques was developed in [11] for Arabic handwritten recognition with improvements in recognition rates.

Using morphological and template matching, an offline interpreter for handwritten Yoruba language recognition was developed in [12].

However, most of the natural language based approaches do not account for variations in dialects for even a language such as Yoruba may support as high as 5 local dialects [13]. In this regard, better models are desired that account for this variation in addition to the utmost task of good feature representation for multi-dialect handwritten data based on experimental evidence.

## 3. Representation Learning and Generative Models

### 3.1 Human Representation of knowledge

We begin this section with the question, "What makes a good Representation"?

Following the definition in [14], we define a good representation as one which facilitates learning features of data easily.

From the human perspective, a good representation is sufficient for projecting a sub-set of useful prior observations onto a posterior memory surface or field. For instance, humans find it easier to recount observations by making a first sequence of observations and assigning a name (or in ML terms a genetic code) to re-occurring samples of observations for subsequent recollection and identification at a later date. In this sense, we may safely say that the representation is "unique", which is another key feature of a good representation. It should be noted here that the uniqueness of a representation follows an automatization process not necessarily defined or inferred consciously by human agents – one interesting definition of a good representation is that it is a uniquely sparse-distributed feature of sequences of data in a hidden markov random field.

### 3.2 Concept of Multi-Dialect Representation

We all speak at least one native (or local) language but do we all have a handwritten representation of what we say in a Second language? The native language is typically referred to as a first language.

In Nigeria, the second language for most citizens is generally English – or more correctly, Nigerian English. Since the quest for handwritten dialects and the challenges that follow demands for a robust and non-restrictive socio-linguistic approach, the "Write-the-Way-You-Hear" (WYH) project, was born [15].

In the WYH project, local handwritten dialects from various ethnic tribes was collated in a questionnaire like form – see Appendix 1. The dialects were transformed to their English equivalent by the contributing speakers interviewed after some prior guides were given.

Basically, the project is about acquiring local handwritten dialects representations using the English letter alphabets. This is the first human representation task developed by the project.

If we assume a basic knowledge of English letters and phonetics, then the following relation may hold:

$$D_{V-n(i)} = P_{ho(i)} \otimes E_g \tag{1}$$

.where,

$D_{V-n(i)}$ = the English language representation of a dialect

$P_{ho(i)}$ = the phonetics as perceived by the writer/speaker

$E_g$ = English alphabets used in the transformation process

$\otimes$ = a transformation function

$i$ = the samples of dialect transformed

Equation (1) shows that the representation of a dialect will be largely dependent on the perceiver i.e. the writer or speaker. This makes human-machine language representations more challenging and interesting.

If space is a concern, the writer may be forced to make adjustments to his or her representations by introducing sparsity terms as:

$$D_{V-n(i)} = \otimes(P_{ho(i)} \otimes A_{s(i)}) \otimes E_g \qquad (2)$$

Where As is probabilistically less than unity.

It must be emphasized that the addition of a word embedding such as in the "Doctor-Nurse" experiments by Meyer & Schvaneveldt in [16] can greatly enhance representational feature learning.

### 3.3 Generative Modelling

#### 3.3.1 Generative Models

Generative models are basically probabilistic models that hypothetically describe a data generating process [17]. They have also been described as data interpreters constructed from a conditional probability distribution [18]. Here, we define them as models for automatically synthesizing stochastic learning units. Generative models play a vital role in modern day task driven ML systems. In this section, we shall describe one very popular generative model- the RBM. We shall also present a modified form of the Rectified Linear Units (ReLU's) used in current deep learning. Finally, we shall introduce a novel and unfamiliar generative model – the Mode Synthesizing Machine (MSM).

#### 3.3.2 The Restricted Boltzmann Machine (RBM)

A generative model such as an RBM may be defined by an Energy function given as [7], [14]:

$$E(x) = -(x^T U x + b^T x) \qquad (3)$$

.where,

$U$ = a weight matrix of model parameters influencing an observation x,

$b^T$ = a transposition bias vector

$x^T$ = a transposition x-vector.

For an Energy function, the joint probability distribution is subsequently given as:

$$P(x) = \frac{\exp(E(x))}{Z} \qquad (4)$$

.where Z is a partition or normalizing function, typically described as:

$$Z = \sum\sum -E(x) \qquad (5)$$

We may rewrite (3) as:

$$E(x) = -(xU + b) \qquad (6)$$

.which is easily recognizable as the familiar representation of most standard neural networks.

Now in real Boltzmann machines, the Energy function, $E(x)$ is typically a summation over the visible and hidden vectors of an observation as:

$$E(x_v, x_h) = -\{(x_v^T W x_h) + (b^T x_v)\} \qquad (7)$$

.where,

$x_v$ = visible units in x

$x_h$ = hidden units in x

$W$ = corresponding weight vectors in $x_v$

Introducing sub-set parameter notation, x is further decomposed into:

$$E(x_v, x_h) = -\{(x_v^T R x_v + x_v^T W x_h) + (x_h^T S x_h) + (b^T x_v + C^T x_h)\} \qquad (8)$$

.where, *R, b, S, C* are additional model parameters constrained on the input vector, x.

Since Boltzmann learning considers maximum likelihood rules, for an independent and identically distributed (i.i.d) data vector of *n* examples, this amounts to maximizing the log-likelihood:

$$\ell(_n) = \log(P(x_v) = \sum_{t=1}^{n} \log(P(X_v^{(t)}) \qquad (9)$$

But Boltzmann machines are parametized not only on the visible units, but on the hidden units as well which is given by:

$$P(x_v^{(t)}) = \sum P(x_v^{(t)}, x_h^{(t)}), \quad \{^{x_v \text{ has priority over } x_h} \tag{10}$$

$$= \sum \frac{1}{Z} \exp\{-E(x_v^{(t)}, x_h^{(t)}))\}$$

Combining (9) and (10) gives an objective function:

$$\ell(_n) = \sum \log \left\{ \frac{1}{Z} \exp\{-E(x_v^{(t)}, x_h^{(t)}))\} \right\} \tag{11}$$

Ideally, we need to analytically solve for Z i.e. obtain a good Z that works well. In practice, this is difficult to achieve - Z is not amenable to analytic solution since it depends on certain free parameters constrained on the visible and hidden data vectors.

Fortunately, we can use the gradient of $\ell(_n)$ with respect to the weight vector, $W$, to improve our objective as [19]:

$$\frac{\partial \ell(_n)}{\partial W} = (x_v, x_h)_{data} - (x_v, x_h)_{model} \tag{12}$$

.which gives a useful learning rule:

$$\Delta_w = \in \{(x_v, x_h)_{data} - (x_v, x_h)_{model}\} \tag{13}$$

.where,

$\in$ = learning rate.

For $T$ visible data vectors, (12) may be rewritten as:

$$\frac{\partial \ell(_n)}{\partial_n} = \frac{1}{T} \sum \{E(x_h, x_{v_t})(\frac{\partial E(x_{v_t}, x_h)}{\partial_n}) - E(x_v, x_h)(\frac{\partial E(x_v, x_h)}{\partial_n}) \tag{14}$$

However, obtaining unbiased samples or approximate estimates of the model is intractable. Practically, one popular way to work around this limitation is by using some sort of Markov Chain Monte Carlo such as Gibbs sampling or Metropolis-Hastings Algorithm [7].

Thus, the model term in (14) is transformed as:

$$E_{v,h} = -E(x_v, x_h)(\frac{\partial E(x_v, x_h)}{\partial_n})$$

$$\approx \frac{1}{M} \sum_{\tilde{x}_{v(m)} \in N} E_{x_h | \tilde{x}_{v(m)}} \left[ (\frac{\partial E(\tilde{x}_{v(m)}, x_h)}{\partial_n} \right] \tag{15}$$

.where,

$M$ = number of parallel markov chains (gibb samples) in the gibb sampler

$\tilde{x}_{v(m)}$ = negative sample

$N$ = the set of negative samples from the RBM data distribution

In an RBM, learning is thus achieved by sampling through the positive and negative phases of data generating distribution in a bid to minimize the cross-entropy and K-L divergence between the model and data generating distribution [14].

### 3.3.3 Rectified Linear Units (ReLU)

Rectified units originally developed in [20] are currently a useful way of learning RBM's in a more expressive. It is an approximation to the Gaussian noise with zero mean and unit variance [19] and may be expressed as:

$$\Re_e^+ = \max(cut-off, x + N(\sim(x), \dagger(x))) \tag{16}$$

.where,

$cut-off = 0$

$N(\sim(x), \dagger(x)))$ = Gaussian noise distribution

$\sim(x)$ = the Gaussian mean – typically 0

$\dagger(x)$ = the Gaussian variance - typically 1

Equation (16) clearly shows that the shape of the representation will be a half-wave of the input data – refer to Figures 1 and 2. However, in certain circumstances, the half-wave version is not at all very useful. For this reasons modifications have been developed to suit diverse applications.

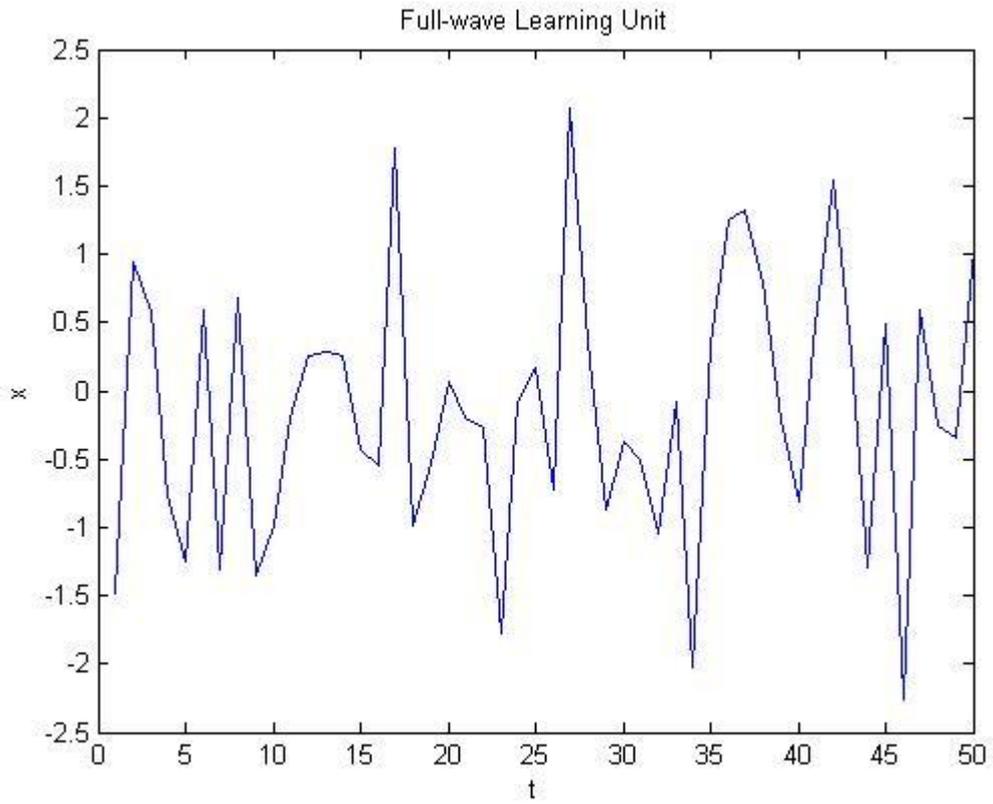

Figure1. A Full-wave Learning Unit before rectification

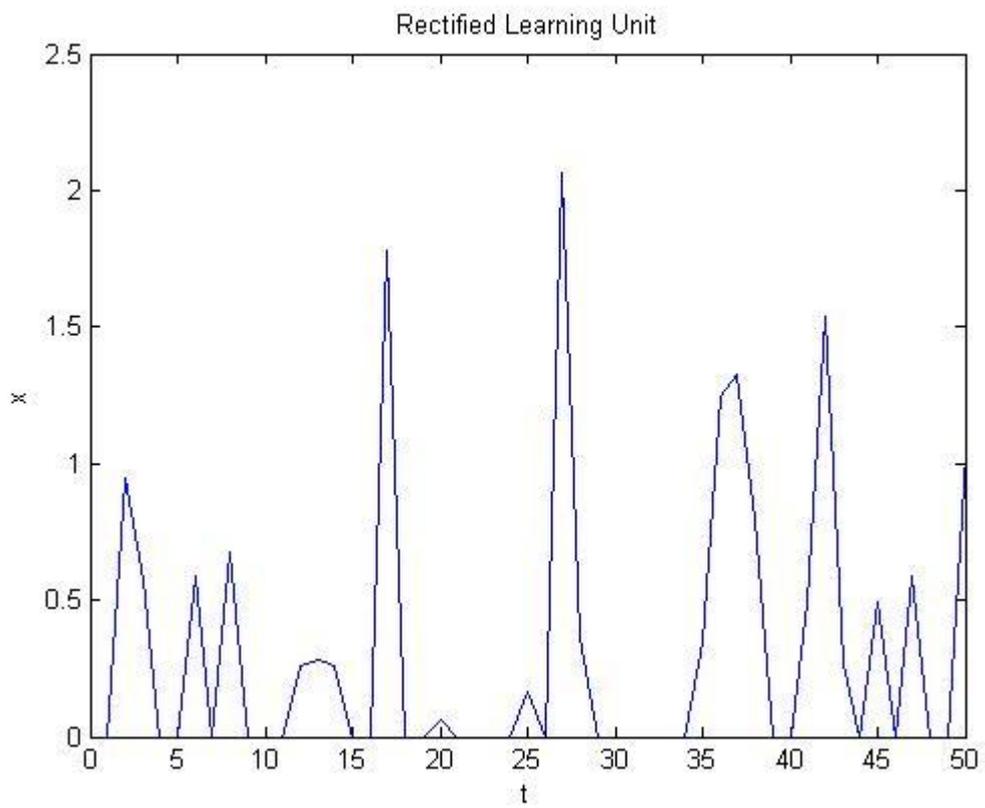

Figure2. A Full-wave Learning Unit after rectification

One approach is to use the means of sample observations, *x*, as a cut-off and replace in (16) as:

$$\Re_e^+ = (\max(\sim(x), x + N(\sim(x), \dagger(x))))^n \qquad (17)$$

where *n* is sparsity exponent – typically in the range of 1-100

This has the effect of averaging over the observation and capturing a sparse set of useful data points (or features in the observation). However, it might not be suitable for all applications so adequate experimenting still has to be carried out.

For this purposes of this study we shall employ (17) as the core-learning unit in the MSM architecture.

### 3.3.4 Mode Synthesizing Machine (MSM)

An MSM is a generative model that tries to capture patterns in unlabeled data by using the mode of a data sequence as a core statistic. It does this basically by first performing a double-linearization on the input data set to obtain the hidden units and then hierarchically synthesizing the modes for each sampled units while performing error minimization through a two-point error update procedure. The implementation algorithm of an MSM is as follows:

1. Initialise errors, first point (Err), second point (err)
2. Set number of epochs:
3. Load data vector Io
4. Compute linear activations

    4.1 Stage1: Compute $I_{o(1)} = 0.2*(1+Io) + 0.5*(1-Io)$

    4.2 Stage2: Compute $I_{o(2)}$ -the learning unit activations using (17)

5. Compute the mean and standard deviation of the input data vector $\sim_o, \dagger_o$
6. *Start Gibbs Sampler, synthesize modes, update errors*

    *for each epoch*

    $\{M_o^*, count\} = M_o(I_{o(2)})$    //Compute the modes of data vector and get the index-count

    *for each row, ro in column, co*

$$m_{rand} = \sim_o + \dagger_o * m_o \begin{cases} m_o \in \cup (0, ro) \\ m_o \in Z \end{cases} \quad \textit{//Generate random units using } \sim_o, \dagger_o$$

$$q_o^* = {}^{ro}P_1 \begin{cases} ro \in Z^+ \\ q_o \in \cup (0, ro); \; qo \in Z^+ \end{cases} \quad \textit{//Generate random permutations}$$

**for each count**

$\quad m_{rand}(count) = M_o^*$     *//Insert- mode of data vector into generative training vectors*

$\quad m_{(1), rand} = m_{rand}(q_o^*, co) \quad \{\forall co$     *//random-permutate the modes in training vector*

$\quad Err = m_{(1), rand} - I_{o(2)}$     *//Compute first-point error*

**end**

$err = | m_{(1), rand} - I_{o(2)} |$ *//compute second-point error*

**end**

*Update Errors and back-propagate by summation*     *// Feedback*

**end**

Thus, we might describe the process as a maximum likelihood each time a sample is generated as the MSM tries to learn maximally the frequencies of each cell or column in the data generating distribution most likely responsible for the visible region of the data. This may be regarded as an extended view of the Maximum-Likelihood Algorithm (MLA).

## 4. Experiments and Results

### 4.1 Experimental Details

Experiments carried out on the test data set compared the RBMs with the MSM based on their reconstruction error. The code for the RBM was adapted from [21] which is based on the mathematical model described in Section 3.3.2 while the MSM was implemented in Matlab ® 7.5 r2007b, based on the algorithm described in Section 3.3.4. With the exception of the Contrastive Divergence (CD) levels and the number of hidden units, default values of learning parameters for the RBM were used. The MSM used a sparsity exponent of 1 for the modified ReLU. Data set for the training have been obtained using questionnaire-like template (see Appendix 1) from native speakers in the South-South region of Nigeria. The Data-set consists of over 1000 handwritten English-letter transformations of local dialects of which 20 have been selected for initial experimentations. Region of interest (ROI) processing have been applied to the selected dataset to extract the handwritten dialects. Experiments were conducted in two parts.

### 4.2 Approach 1

For the first experiment we generate representations for a single handwritten dialect to validate MLA proof of concept for both the RBM and MSM. We perform contrastive divergence (CD) at CD-1 and CD-500 levels and note our observations. The number of hidden units chosen for the RBM is 500.

### 4.3 Approach 2

For the second test, we generate mixed representations for multiple dialects concatenated through a recurrent sequential process (not shown) to validate MLA proof of concept and study the mixed signal response for both the RBM and MSM. Contrastive divergences at the CD-1 and CD-500 levels are utilized for training and reconstructions. The number of hidden units chosen for the RBM is 500.

### 4.4 Results

The original image for Approach 1 is shown in Figure 3 while the reconstructed images using Approach 1 at the CD-1 level is as shown in Figures 4 and 5 for RBM and MSM respectively. The reconstruction errors for Approach 1 and Approach 2 are given in Tables 1 and 2 respectively.

From the results it is obvious that the MSM fared much better than the RBM. There is no guarantee that the RBM or MSM will improve much well than its current state even at much higher CD-levels.

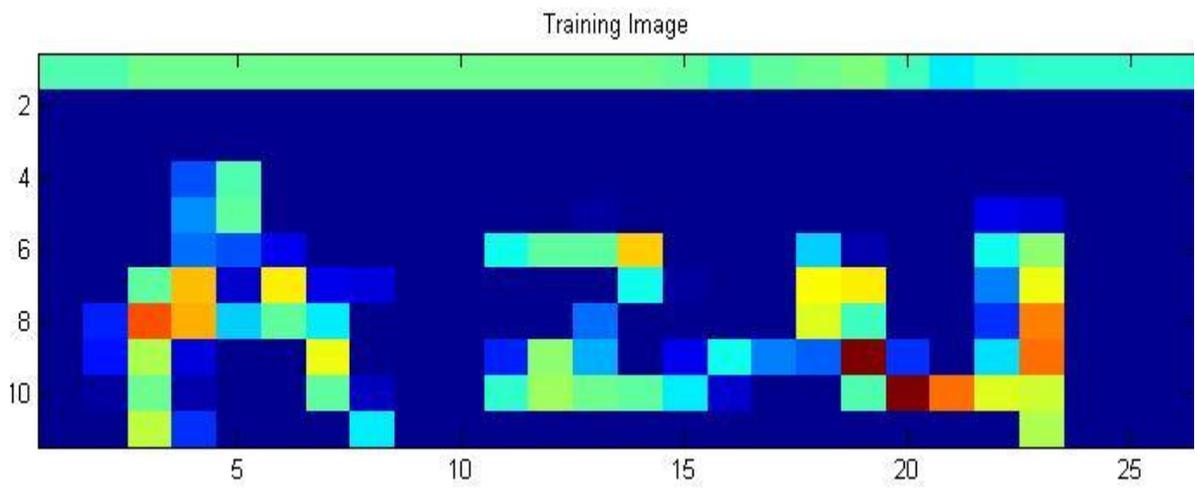

Fig.1 Image of the handwritten dialect *Azu* after ROI processing

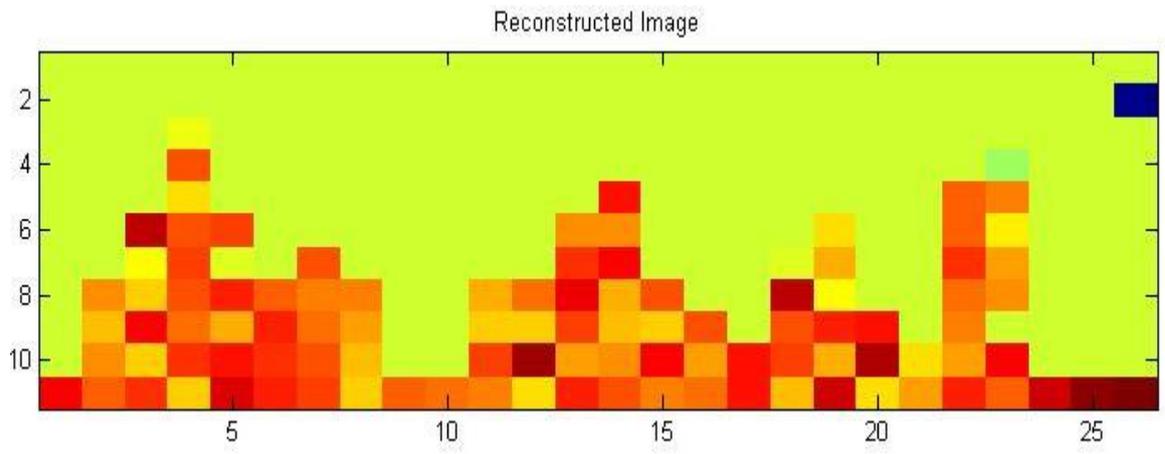

Fig.2 A generative model for the handwritten dialect *Azu* at CD-1 using MSM. Notice the sharp contrast in the images. However it is still easy to deduce the shape of the generative model from the shape of the original image.

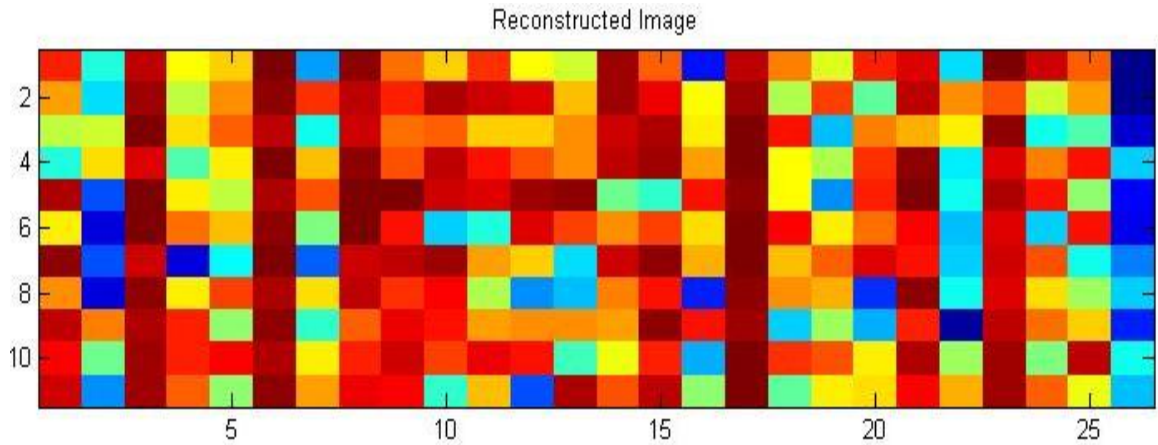

Fig.3 A generative model for the handwritten dialect *Azu* at CD-1 using RBM. Notice the sharp contrast in the images. However it is not easy to deduce the shape of the generative model from the shape of the original image

Table1. Reconstruction accuracies for Approach 1 after a minimum of 2 trials

| Model | Reconstruction error (CD-1) | Reconstruction error (CD-500) |
|---|---|---|
| RBM | 34.000 | 6.1000 |
| MSM | 0.2076 | 0.1955 |

Table2. Reconstruction accuracies for Approach 2 after a minimum of 2 trials

| Model | Reconstruction accuracy (CD-1) | Reconstruction accuracy (CD-500) |
|---|---|---|
| RBM | 1003.3 | 57.000 |
| MSM | 6.1473 | 6.2072 |

## 5. Conclusions

This paper has described a new concept for multi-dialect handwritten representation and a generative model using two-point error update suitable for learning these rich handwritten representations. We have additionally described a double linear learning unit using a first-order modified Rectified Linear Unit (ReLU) which gives promising results for the MSM. From our experiments MSM fared much better than the RBM at the CD-1 and CD-500 level. The MSM data representations may be improved further by using tensor analyzers (TA) and recurrent Long-Short-Term Memory (rLSTM) sparse distributed representations.

# Appendix 1

# A WYW Template for Handwritten Local Dialect Representation

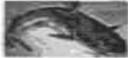